%% file: AnonymousSubmission2027.tex
\documentclass[letterpaper]{article} 
\usepackage{lmodern}
\usepackage[preprint]{aaai2027}  
\usepackage[hyphens]{url}  
\usepackage{graphicx} 
\usepackage{natbib}  
\usepackage{caption} 
\definecolor{linkblue}{RGB}{0,90,160}

\usepackage[
    colorlinks=true,
    urlcolor=linkblue,
    citecolor=black,
    linkcolor=black
]{hyperref}
\usepackage{amsmath}
\usepackage{array}
\usepackage{multirow}
\usepackage[table]{xcolor}
\definecolor{oursrowblue}{RGB}{220,230,241}
\usepackage{amssymb}

\usepackage{booktabs}
\usepackage[most]{tcolorbox}
\usepackage{xcolor}

\definecolor{questionbg}{HTML}{F3F7F5}
\definecolor{questionaccent}{HTML}{168A68}
\definecolor{questiontext}{HTML}{1F2933}
\newtcolorbox{researchquestion}{
    enhanced,
    breakable,
    colback=black!4,
    colframe=black!4,
    coltext=black!85,
    boxrule=0pt,
    arc=3pt,
    left=14pt,
    right=14pt,
    top=6pt,
    bottom=6pt,
    before skip=6pt,
    after skip=6pt
}

\title{Beyond Pixels: From Video Priors to 4D Worlds}
\author{
    Zihao Liu\textsuperscript{\rm 1},
    Xiaolong Shen\textsuperscript{\rm 1},
    Zhenglin Zhou\textsuperscript{\rm 1},
    Ruijie Quan\textsuperscript{\rm 1},
    Yi Yang\textsuperscript{\rm 1}\corresponding
}
\affiliations{
    \textsuperscript{\rm 1}ReLER, CCAI, Zhejiang University\\

    Project Page: \url{https://hayd-zju.github.io/Beyond-Pixels/}
}

\makeatletter
\g@addto@macro\@maketitle{%
  \vspace{-60pt}
  \begin{center}
  \captionsetup{font=small,skip=2pt}
  \includegraphics[width=1\textwidth,height=0.55\textheight,keepaspectratio]{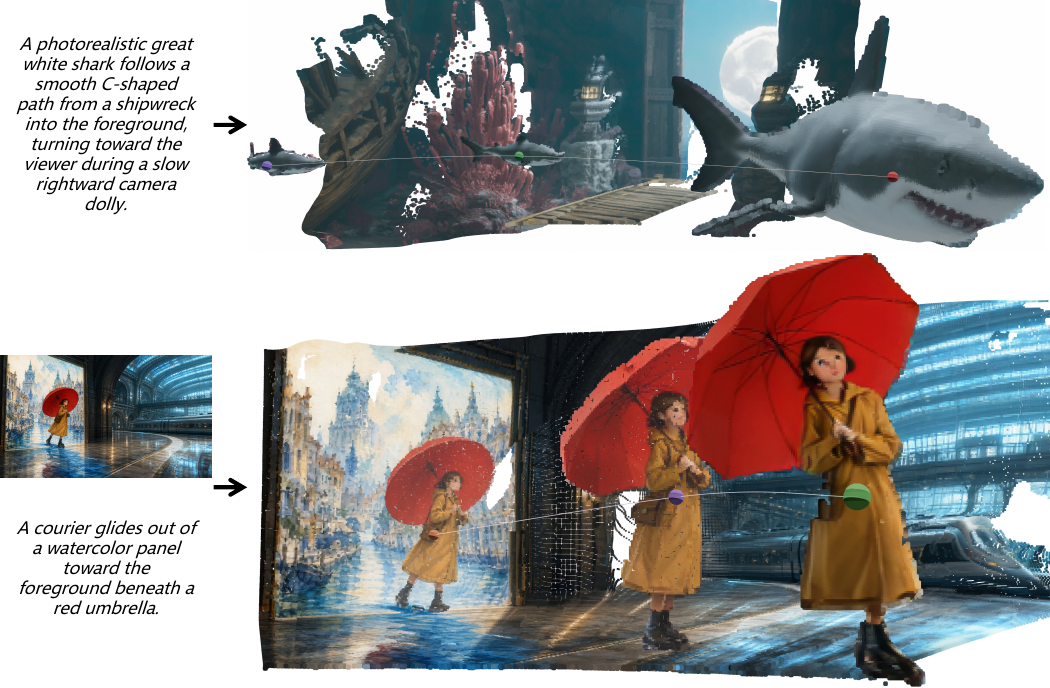}\par
  \captionof{figure}{\textbf{Latent-to-4D} enables unified text- and image-conditioned 4D
    generation. It turns video-model latents directly into dynamic geometry
    and motion through a shared pathway, without task-specific adaptation.}
  \label{fig:task_diversity}
  \end{center}
}
\makeatother

\begin{document}

\maketitle

\input{sections/0_abstract}
\input{sections/1_intro}
\input{sections/2_related_work}
\input{sections/3_method}
\input{sections/4_exp}

\input{sections/5_conclusion}

\bibliography{aaai2027}

\end{document}

%% file: sections/0_abstract.tex
\begin{abstract}
4D generation synthesizes dynamic 3D scenes from conditions such as text or
images. Existing methods either reconstruct generated RGB videos with a
separate 4D model or adapt a particular video generator to predict geometry
directly. The former suffers from distribution mismatch and error propagation, whereas the latter ties 4D prediction to a specific generator and may require retraining when the generator or conditioning regime changes. We ask whether the final denoised latents of video models
that share a variational autoencoder (VAE) can instead provide a reusable
interface to explicit 4D prediction. Building on this insight,
we introduce \emph{direct latent-to-4D
generation} and instantiate it as \textbf{Latent-to-4D}, which bypasses RGB by
aligning a video latent with the token grid of a pretrained 4D decoder and
refining it through frame-wise and global spatiotemporal attention. Trained on roughly 1K existing reconstruction clips, a single checkpoint transfers unchanged across multiple video diffusion transformers within the same VAE family. On Text4D-200 and I4D-200, Latent-to-4D surpasses matched same-latent Wan+4RC cascades in projection-based DINO-F1 by 2.88--3.45 and 5.81 points, respectively, while also being preferred by human raters for geometry, temporal stability, and overall quality. 
\end{abstract}

%% file: sections/1_intro.tex
\section{Introduction}
\label{sec:intro}

4D generation aims to synthesize dynamic 3D scenes with explicit geometry and motion from intuitive conditions such as text or images. Such outputs can be viewed and manipulated from novel viewpoints, making them valuable for virtual production, immersive VR/AR, simulation, navigation, and embodied intelligence. Recent advances in video generation and feed-forward 4D reconstruction have made this goal increasingly practical, motivating approaches that combine their complementary generative and geometric capabilities.

Recent 4D generation methods combine these capabilities through two broad paradigms. The first paradigm is \textbf{generate-then-reconstruct}~\cite{diffusion4d, 4diffusion, cat4d}: They first synthesize multiview or temporal RGB observations, which are then converted into an explicit dynamic scene by a separate reconstruction model. The second is \textbf{integrated feed-forward generation}~\cite{diff4splat,4dnex,worldreel}, which makes geometry a native output of the generative process through either direct prediction of explicit 4D representations or joint RGB--geometry video modeling. Collectively, these approaches demonstrate the value of video priors for efficient 4D generation. 

Despite this progress, the two paradigms expose a fundamental trade-off in how video priors are transferred to 4D. Generate-then-reconstruct preserves generator modularity, but its RGB interface places a separately trained reconstructor, typically trained on much narrower reconstruction data, between the video prior and the 4D output. Open-domain content and temporal artifacts may therefore propagate into unstable geometry. Integrated feed-forward methods transfer video priors more directly, but typically bind 4D prediction to a particular generator or conditioning regime, so switching to another pretrained video model may require new geometry-supervised training. This trade-off is amplified by the scarcity of 4D supervision relative to large-scale video data. The key challenge is therefore to establish a reusable interface that bypasses generated RGB while allowing one geometry-supervised pathway to serve multiple compatible video generators.

A natural candidate for such an interface is the VAE latent space shared by compatible video generators. Although their DiT backbones and conditioning regimes may differ, models that share the same VAE checkpoint, latent normalization, layout, and compression convention produce final denoised latents in a common representation space. Because this representation lies upstream of RGB decoding, it provides a natural point at which a single geometry-supervised 4D pathway could access the appearance, motion, and conditioning information produced by different compatible DiTs (Fig.~\ref{fig:task_diversity}). This observation motivates our central question: 
\begin{researchquestion}
\centering
\itshape
Can final denoised video latents serve as a reusable interface to explicit 4D prediction across compatible DiTs?
\end{researchquestion}

\begin{figure}[t]
    \centering
    \includegraphics[width=\columnwidth]{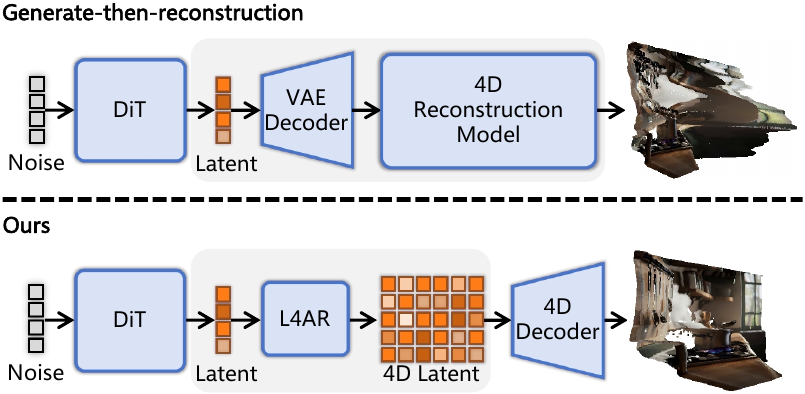}
    \caption{\textbf{Comparison of video-to-4D interfaces.}
    Previous methods decode video latents to RGB before reconstruction, whereas ours maps them directly to 4D through L4AR.}
    \label{fig:intro_comparison}
    \vspace{-10pt}
\end{figure}

To realize this interface, we propose \textbf{Latent-to-4D}, a direct 4D generation framework that connects a video model's final denoised latent to a 4D decoding hierarchy without passing through generated RGB (Fig.~\ref{fig:intro_comparison}).  Bridging these representations is non-trivial because their spatiotemporal grids and feature spaces are not aligned. Our Latent-to-4D Alignment and Refinement (\textbf{L4AR}) network aligns the video latent with the 4D token grid and refines it using frame-wise and global spatiotemporal context, after which a decoder initialized from a pretrained reconstructor predicts cameras and dynamic world-space geometry.  During training, a frozen VAE encodes roughly 1K existing reconstruction clips, and only the alignment module, lightweight refinement updates, and prediction heads are optimized from their 4D annotations. The video generators, VAE, and original Transformer weights remain frozen.
At inference, the observed-video latent is replaced by the final denoised latent of a compatible DiT in the same VAE space, allowing the trained pathway to produce 4D outputs without further tuning.

We evaluate Latent-to-4D on \textbf{Text4D-200} and \textbf{I4D-200}, two generated-latent evaluation suites for text- and image-conditioned 4D generation. A single checkpoint operates unchanged across two text-to-video DiTs and one image-to-video DiT sharing the same VAE, while remaining compatible with their upstream controls. In controlled same-latent comparisons, Latent-to-4D surpasses matched Wan+4RC cascades in projection-based DINO-F1 by 2.88--3.45 points on Text4D-200 and 5.81 points on I4D-200, and is preferred in multi-view human evaluation for geometric plausibility, completeness, and temporal stability. These results support shared video latents as an effective and reusable interface within the evaluated common-VAE setting.

Our contributions are threefold:
\begin{itemize}
    \item Conceptually, we formulate direct latent-to-4D generation by treating final denoised VAE latents as a reusable interface between compatible video generators and explicit 4D prediction. This formulation bypasses generated RGB and decouples 4D supervision from a particular generator or conditioning regime.
    
    \item Technically, we introduce \textbf{Latent-to-4D}, whose
    L4AR network aligns and refines video latents for structured 4D decoding.
    Trained on roughly 1K existing reconstruction clips, a single checkpoint transfers unchanged across multiple video generators sharing the same VAE.
    
    \item Empirically, we evaluate a single Latent-to-4D checkpoint across multiple video generators sharing the same VAE on Text4D-200 and I4D-200. Latent-to-4D achieves higher projection-based DINO-F1 than matched generate-then-reconstruct baselines, while human evaluation favors its outputs for geometric plausibility, completeness, and temporal stability.
    
\end{itemize}

%% file: sections/2_related_work.tex
\section{Related Work}

\noindent
\textbf{Video Generation.}
Modern video generators learn rich appearance and motion distributions and
support text, image, camera, trajectory, and animation controls
\cite{videocrafter2,cogvideox,hunyuanvideo,wan,stablevideodiffusion,
i2vgenxl,dynamicrafter,motionctrl,cameractrl,recammaster,wanmove,wananimate}.
Their representations have also been repurposed for depth, point maps, and
dynamic geometry~\cite{marigold,depthcrafter,geometrycrafter,geo4d,sora3r,
motioncrafter,genception}, with VIST3A connecting a video VAE to a static 3D
reconstructor~\cite{vist3a}. These approaches typically convert a generator
into a task-specific perception model or align one selected model pair. We
instead keep compatible conditional DiTs and their shared VAE frozen and learn
one interface from the final denoised VAE latent to a dynamic 4D decoder,
preserving their generation and control capabilities.

\noindent
\textbf{4D Reconstruction.}
Feed-forward 4D reconstructors provide reusable geometric reasoning by
predicting dynamic point maps, scene flow, trajectories, or dense motion from
RGB videos
\cite{st4rtrack,dynamicpointmaps,traceanything,any4d,vdpm}. Query-based models
such as D4RT and 4RC recover cameras and geometry across viewpoint and time,
while related models such as $\pi^3$ estimate cameras and dense point maps from
image sets~\cite{d4rt,4rc,pi3}. Their pipelines nevertheless begin with RGB
encoders trained on reconstruction data; when appended to a generator, these
encoders must interpret potentially out-of-distribution generated frames. We
retain the pretrained 4D reconstruction hierarchy as the decoder and replace
this RGB interface with an aligned final denoised VAE latent, thereby using the
video model itself as the encoder.

\noindent
\textbf{4D Generation.}
Existing 4D generation methods combine generative priors with dynamic scene
representations through three main routes. Earlier text- and image-guided
methods optimize an object-centric dynamic NeRF or Gaussian representation
separately for each output
\cite{mav3d,animate124,4dfy,dreamin4d,alignyourgaussians,comp4d,
dreamscene4d}. Generate-then-reconstruct methods, including Diffusion4D,
4Diffusion, and CAT4D, synthesize multiview or temporal RGB observations before
recovering an explicit dynamic scene~\cite{diffusion4d,4diffusion,cat4d}.
Feed-forward alternatives such as 4DNeX and Diff4Splat adapt video generators
to predict dynamic point clouds or Gaussians for selected image-conditioned
settings~\cite{4dnex,diff4splat}; WorldReel jointly predicts RGB and geometry,
while WorldForge adds inference-time camera control
\cite{worldreel,worldforge}. Latent-to-4D instead connects a frozen video model
to a pretrained 4D decoder directly in latent space, using the former as the
encoder: it neither optimizes each scene nor retrains each generator, and does
not use generated RGB as the geometric input.

%% file: sections/3_method.tex
\section{Method}
\label{sec:method}

\begin{figure*}[t]
    \centering
    \includegraphics[width=0.85\textwidth]{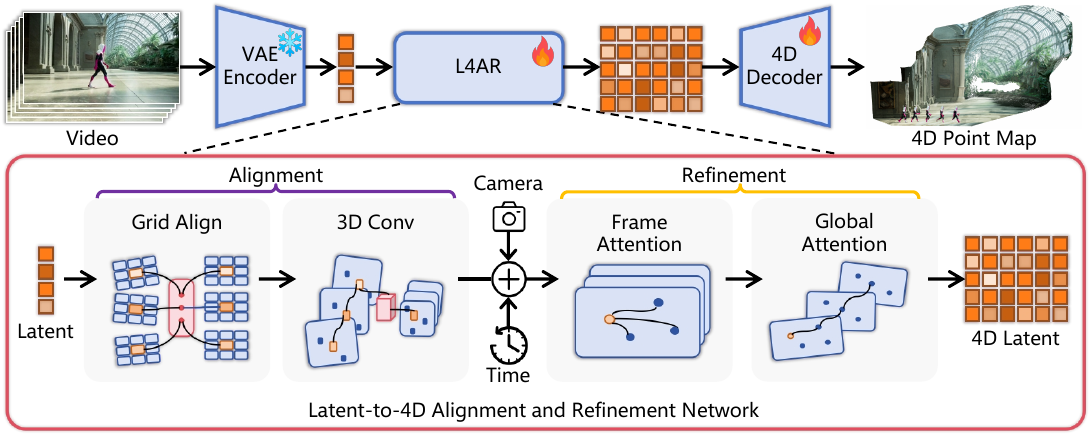}
    \caption{\textbf{Latent-to-4D training pipeline.}
     A frozen video VAE encodes an observed video into a VAE-space latent. L4AR aligns the latent grid through a learned 3D convolution, reuses frozen camera and time tokens, and refines the representation through alternating frame-wise and global attention. The 4D decoder predicts cameras and dynamic world-space geometry.}
     \vspace{-10pt}
    \label{fig:method_overview}
\end{figure*}

Latent-to-4D instantiates direct latent-to-4D generation by replacing an RGB encoder with the video model's VAE-space representation. As in Fig.~\ref{fig:method_overview}, a frozen VAE provides observed-video latents during training, whereas a compatible frozen DiT provides final denoised latents during generation. The same pathway aligns either source, refines it with pretrained 4D latent, and decodes cameras and dynamic geometry.

\subsection{Problem Formulation}
\label{sec:problem_formulation}

We formulate direct latent-to-4D generation as a cross-representation mapping
from a video VAE latent space to the structured token space of a pretrained 4D
reconstructor. Let $(E_{\mathrm v},D_{\mathrm v})$ denote the spatiotemporal
VAE of a video generator, and let $\mathcal Z_{\mathrm v}$ denote its latent
space. A video latent $\mathbf{z}_{\mathrm v}\in\mathcal Z_{\mathrm v}$ is
used to encode the appearance and motion information required for video
decoding. By contrast, the pretrained 4D hierarchy operates on a structured
token space $\mathcal Z_{\mathrm{4D}}$, whose elements
$\mathbf{Q}\in\mathcal Z_{\mathrm{4D}}$ support camera and dynamic-geometry
prediction. Although both representations are spatiotemporal, they differ in
temporal resolution, spatial grid, and feature dimension and are therefore not
directly interchangeable.

The central problem is to bridge these two representation spaces without
passing through RGB. Specifically, we seek an alignment map
\begin{equation}
\mathcal A_\phi:
\mathcal Z_{\mathrm v}\rightarrow\mathcal Z_{\mathrm{4D}},
\qquad
\mathbf{Q}^{(0)}=\mathcal A_\phi(\mathbf{z}_{\mathrm v}),
\label{eq:cross_representation_mapping}
\end{equation}
such that the aligned tokens $\mathbf{Q}^{(0)}$ can be refined and decoded
into a discrete 4D scene
\begin{equation}
\mathcal Y
=\left\{(\mathbf{C}_t,\mathbf{P}_t)\right\}_{t=1}^{T},
\qquad
\mathbf{P}_t\in\mathbb R^{H\times W\times 3},
\label{eq:4d_target}
\end{equation}
where $\mathbf{C}_t$ denotes the camera at time $t$ and $\mathbf{P}_t$ is a
dense point map expressed in a shared world coordinate system. This
representation captures both time-varying geometry and the camera trajectory
required to render the scene from novel viewpoints.
A conventional generate-then-reconstruct pipeline predicts
\begin{equation}
\widehat{\mathcal Y}_{\mathrm{rgb}}
=R\!\left(D_{\mathrm v}(\mathbf{z}_{\mathrm v})\right),
\label{eq:rgb_interface}
\end{equation}
where $D_{\mathrm v}$ first decodes the video latent into RGB frames and
an independently trained reconstructor $R$ subsequently lifts those frames
into 4D. This pipeline introduces an RGB representation boundary through
which generation and decoding artifacts can propagate to the geometry
prediction. Our objective is instead to learn a direct mapping
$\mathcal Z_{\mathrm v}\rightarrow\mathcal Z_{\mathrm{4D}}
\rightarrow\mathcal Y$ that bypasses both RGB decoding and subsequent RGB
re-encoding.

\subsection{Shared-Latent Interface across Compatible DiTs}
\label{sec:generative_4d_reconstruction}

We next exploit the video VAE space as a shared input interface across
compatible DiTs. Given an observed video tensor $\mathbf{V}$, the VAE encoder
provides the posterior-mean latent
$\mathbf{z}^{\mathrm{obs}}
=\mu(E_{\mathrm v}(\mathbf{V}))\in\mathcal Z_{\mathrm v}$.
Given a condition $c$ and sampled noise $\boldsymbol{\epsilon}$, a compatible
video DiT $G_\theta$ produces the final denoised latent
$\mathbf{z}^{\mathrm{gen}}
=G_\theta(c,\boldsymbol{\epsilon})\in\mathcal Z_{\mathrm v}$
immediately before VAE decoding. Both latent sources follow the same VAE
scaling, tensor layout, and compression convention, although their
distributions need not coincide.

For either source
$\mathbf{z}\in\{\mathbf{z}^{\mathrm{obs}},\mathbf{z}^{\mathrm{gen}}\}$,
Latent-to-4D factorizes the direct mapping into alignment, spatiotemporal
refinement, and structured 4D decoding:
\begin{equation}
\widehat{\mathcal Y}
=\mathcal D_\omega\!\left(
\mathcal H_{\psi,\Delta\psi}
\bigl(\mathcal A_\phi(\mathbf{z});\mathbf{S}\bigr)
\right).
\label{eq:generative_4d_pipeline}
\end{equation}
The Alignment Module $\mathcal A_\phi$ maps the VAE latent tensor to the
spatiotemporal token grid expected by the pretrained 4D reconstruction
hierarchy. The Spatiotemporal Refinement Module
$\mathcal H_{\psi,\Delta\psi}$ then transforms the aligned tokens into
contextual 4D latent using frozen pretrained weights $\psi$ and lightweight
trainable updates $\Delta\psi$; $\mathbf{S}$ denotes the frozen camera and time
tokens. Finally, the 4D Decoder $\mathcal D_\omega$ predicts per-frame cameras
and dynamic world-space geometry.

During training, the pathway receives $\mathbf{z}^{\mathrm{obs}}$ from videos
with 4D annotations, and supervision updates the Alignment Module, the
lightweight Refinement updates, and the geometry and camera heads. The video
DiT is neither executed nor optimized. At inference,
$\mathbf{z}^{\mathrm{obs}}$ is replaced by $\mathbf{z}^{\mathrm{gen}}$, while
the complete downstream pathway remains unchanged. Because the condition has
already been realized in the generated latent, no condition-specific 4D
branch or explicit task identifier is required.

Compatibility requires the video DiTs to share the VAE checkpoint, latent
normalization, tensor layout, compression convention, and a supported latent
shape. Within this common-VAE boundary, their DiT architectures and
conditioning regimes may differ. In the Experiments section, we evaluate a
single Latent-to-4D checkpoint across two text-to-video DiTs and one
image-to-video DiT, while the ``Sensitivity to DiT-Derived Residuals'' examines its
sensitivity to a controlled DiT-derived residual component.

\subsection{L4AR for Structured 4D Decoding}
\label{sec:latent_to_4d_alignment}

\noindent
\textbf{Alignment Module.}
The VAE latent cannot be consumed directly by the pretrained 4D hierarchy
because the two representations differ in temporal resolution, spatial grid,
and feature dimension. We first apply a fixed trilinear resampling operator
$\mathcal R$ to match the required spatiotemporal resolution. A learned 3D
convolution $\mathcal S_\phi$ then aggregates local spatiotemporal
neighborhoods and projects the latent channels to the feature dimension
expected by the 4D hierarchy. After flattening the spatial dimensions, the
aligned tokens are
\begin{equation}
\mathbf{Q}^{(0)}
=\mathcal A_\phi(\mathbf{z})
=\operatorname{Flatten}\!\left(
\mathcal S_\phi\!\left(\mathcal R(\mathbf{z})\right)
\right)
\in\mathbb R^{T\times M\times d},
\label{eq:latent_alignment}
\end{equation}
where $T$ is the number of frames, $M$ is the number of spatial tokens per
frame, and $d$ is the token dimension. The 3D convolution performs a shared
local conversion across the latent grid, allowing neighboring appearance and
motion evidence to be aggregated before global reasoning.

\begin{figure*}[t]
    \centering
    \includegraphics[width=\linewidth]{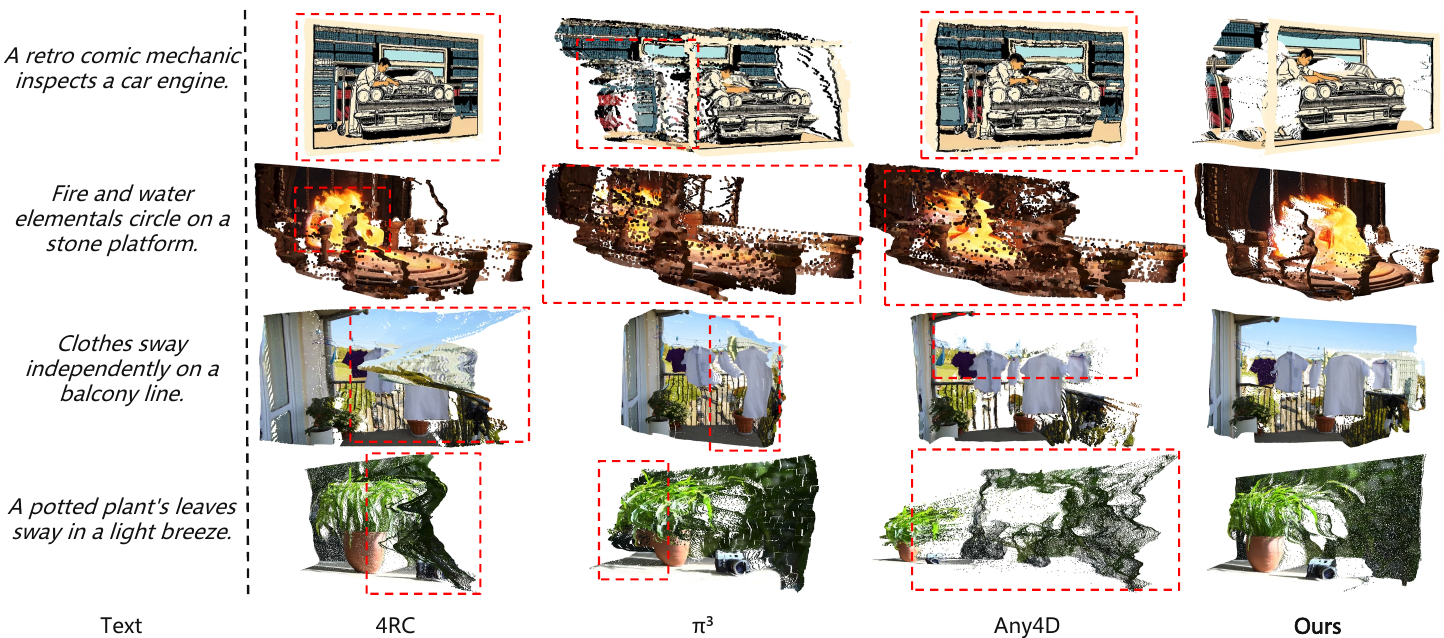}
    \caption{\textbf{Text-to-4D comparison.}
    Texts are abbreviated for display. RGB baselines reconstruct the decoded video, whereas Ours consumes its terminal latent directly.}
    \label{fig:text4d_qualitative}
    \vspace{-10pt}
\end{figure*}

\noindent
\textbf{Spatiotemporal Refinement Module.}
Although alignment produces a compatible token grid, its local receptive
field does not capture long-range correspondence, viewpoint changes, or
motion across a sequence. We therefore introduce a hierarchical refinement
module that combines frame-wise and global spatiotemporal self-attention.
Given tokens of shape $(B,T,M,d)$, frame-wise attention reshapes them to
$(BT,M,d)$ and consolidates spatial structure within each
frame, whereas global attention operates on $(B,TM,d)$ to exchange
information across all spatial locations and time steps. The hierarchy first
establishes per-frame structure through frame-wise attention and then
alternates the two attention modes, allowing global blocks to propagate
correspondence, viewpoint, and motion evidence while interleaved frame-wise
blocks retain detailed spatial structure.

We collect intermediate representations at multiple depths of the hierarchy.
At each selected depth, the most recent frame-wise and global features are
concatenated to form a multi-level representation containing both spatial
detail and sequence-level context. This representation provides the 4D
Decoder with complementary intra-frame geometry and cross-frame motion cues.

\noindent
\textbf{4D Decoder.}
The 4D Decoder converts the refined feature hierarchy into explicit dynamic
scene structure. For each frame $t$ and pixel $u$, a multi-level geometry head
predicts a depth value $\widehat d_t(u)$ and a world-space ray represented by
origin $\widehat{\mathbf{o}}_t$ and unit direction
$\widehat{\mathbf{r}}_t(u)$, together with their confidence values. In
parallel, a camera head predicts $\widehat{\mathbf{C}}_t$ through a 9D
pose--field-of-view parameterization. Following the decoder's ray
parameterization, the corresponding world-space point is recovered as
\begin{equation}
\widehat{\mathbf{P}}_t(u)
=\widehat{\mathbf{o}}_t
+\widehat d_t(u)\widehat{\mathbf{r}}_t(u).
\label{eq:world_point_decode}
\end{equation}
The predicted cameras and dynamic point maps form
$\widehat{\mathcal Y}$, which can be reprojected across viewpoints and time
rather than remaining restricted to the input video observations.

\noindent
\textbf{Training Objective.}
Given an annotated sequence $(\mathbf{V},\mathcal Y)$, the frozen VAE produces
$\mathbf{z}^{\mathrm{obs}}$, and the latent-to-4D pathway is supervised by
\begin{equation}
\mathcal L
=\mathcal L_{\mathrm{unc}}
+\mathcal L_{\mathrm{cam}}
+\mathcal L_{\mathrm{geom}}.
\label{eq:training_objective}
\end{equation}
The uncertainty-aware term $\mathcal L_{\mathrm{unc}}$ contains
confidence-weighted depth, depth-gradient, and world-ray losses.
$\mathcal L_{\mathrm{cam}}$ supervises camera translation, rotation, and
field of view, while $\mathcal L_{\mathrm{geom}}$ combines metric depth and
ray supervision with mean and tail errors on world-space points and a
surface-normal loss. Together, these objectives train the Alignment Module,
the lightweight Refinement updates, and the geometry and camera heads from
metric 4D annotations.

Throughout training, the video generators, VAE, original Transformer weights,
camera and time tokens, motion decoder, and tracking head remain frozen. We
progressively activate the trainable components to avoid destabilizing the
pretrained geometric prior. Dataset composition, LoRA configuration, and
stage-wise training budgets are provided in the
Experimental Setup subsection and the Appendix.

%% file: sections/4_exp.tex
\section{Experiments}
\label{sec:experiments}

\renewcommand{\topfraction}{0.95}
\renewcommand{\bottomfraction}{0.45}
\renewcommand{\dbltopfraction}{0.95}
\renewcommand{\textfraction}{0.04}
\renewcommand{\floatpagefraction}{0.90}
\renewcommand{\dblfloatpagefraction}{0.90}
\setcounter{topnumber}{4}
\setcounter{bottomnumber}{4}
\setcounter{totalnumber}{8}
\setcounter{dbltopnumber}{3}
\setlength{\textfloatsep}{6pt plus 1pt minus 2pt}
\setlength{\dbltextfloatsep}{6pt plus 1pt minus 2pt}
\setlength{\floatsep}{5pt plus 1pt minus 2pt}
\setlength{\dblfloatsep}{5pt plus 1pt minus 2pt}

\begin{table}[t]
\centering
\fontsize{6.4}{6.7}\selectfont
\setlength{\tabcolsep}{1.15pt}
\renewcommand{\arraystretch}{0.84}

\begin{tabular}{@{}>{\raggedright\arraybackslash}p{0.40\columnwidth}
*{5}{>{\raggedleft\arraybackslash}p{0.11\columnwidth}}@{}}
\toprule
Method
& \shortstack{Text\\CLIP $\uparrow$}
& \shortstack{RGB-ref.\\CLIP-I $\uparrow$}
& \shortstack{DINO\\global $\uparrow$}
& \shortstack{DINO\\match $\uparrow$}
& \shortstack{DINO\\F1 $\uparrow$} \\
\midrule
\multicolumn{6}{@{}l}{\textcolor{black!60}{\textit{Text-to-4D}}} \\
CogVideoX-5B + 4RC
& 26.887 & \textbf{75.33} & 42.71 & 53.47 & 53.27 \\
CogVideoX-5B + $\pi^3$
& 26.293 & 74.84 & 38.74 & 50.29 & 49.72 \\
CogVideoX-5B + Any4D
& 25.414 & 72.66 & 29.46 & 45.97 & 45.12 \\
Wan2.1-14B + 4RC
& 28.116 & 71.20 & 42.45 & 54.31 & 53.56 \\
Wan2.1-14B + $\pi^3$
& 26.594 & 68.70 & 34.79 & 48.69 & 47.50 \\
Wan2.1-14B + Any4D
& 26.261 & 67.56 & 29.97 & 46.80 & 45.55 \\
Wan2.1-1.3B + 4RC
& 27.829 & 71.34 & 43.30 & 54.97 & 54.21 \\
Wan2.1-1.3B + $\pi^3$
& 27.034 & 70.23 & 38.51 & 51.23 & 50.10 \\
Wan2.1-1.3B + Any4D
& 26.326 & 68.19 & 31.00 & 47.26 & 46.06 \\
\cmidrule(l){1-6}
\rowcolor{oursrowblue}
\textbf{Ours (Wan2.1-14B)}
& \textbf{28.544} & 72.24 & 45.43
& 57.52 & 57.01 \\
\rowcolor{oursrowblue}
\textbf{Ours (Wan2.1-1.3B)}
& 28.434 & 72.32 & \textbf{46.02}
& \textbf{57.64} & \textbf{57.09} \\
\midrule
\multicolumn{6}{@{}l}{\textcolor{black!60}{\textit{Image-to-4D}}} \\
4DNeX
& 22.844 & 61.11 & 11.31 & 30.53 & 28.33 \\
Wan2.2-I2V-A14B + 4RC
& 26.340 & 70.55 & 47.83 & 56.25 & 55.79 \\
Wan2.2-I2V-A14B + $\pi^3$
& 24.678 & 66.65 & 33.50 & 45.08 & 43.82 \\
Wan2.2-I2V-A14B + Any4D
& 24.362 & 65.30 & 27.21 & 43.54 & 42.17 \\
\cmidrule(l){1-6}
\rowcolor{oursrowblue}
\textbf{Ours}
& \textbf{27.340} & \textbf{72.87} & \textbf{54.85}
& \textbf{61.85} & \textbf{61.60} \\
\bottomrule
\end{tabular}
\caption{\textbf{4D generation on Text4D-200 and I4D-200.}
Two-view projection scores (${\times}100$; higher is better); Ours and
matched Wan cascades share the same generated latent.}
\label{tab:generated4d_main}
\end{table}

\begin{figure*}[t]
    \centering
    \vspace{-5pt}
    \includegraphics[width=\linewidth]{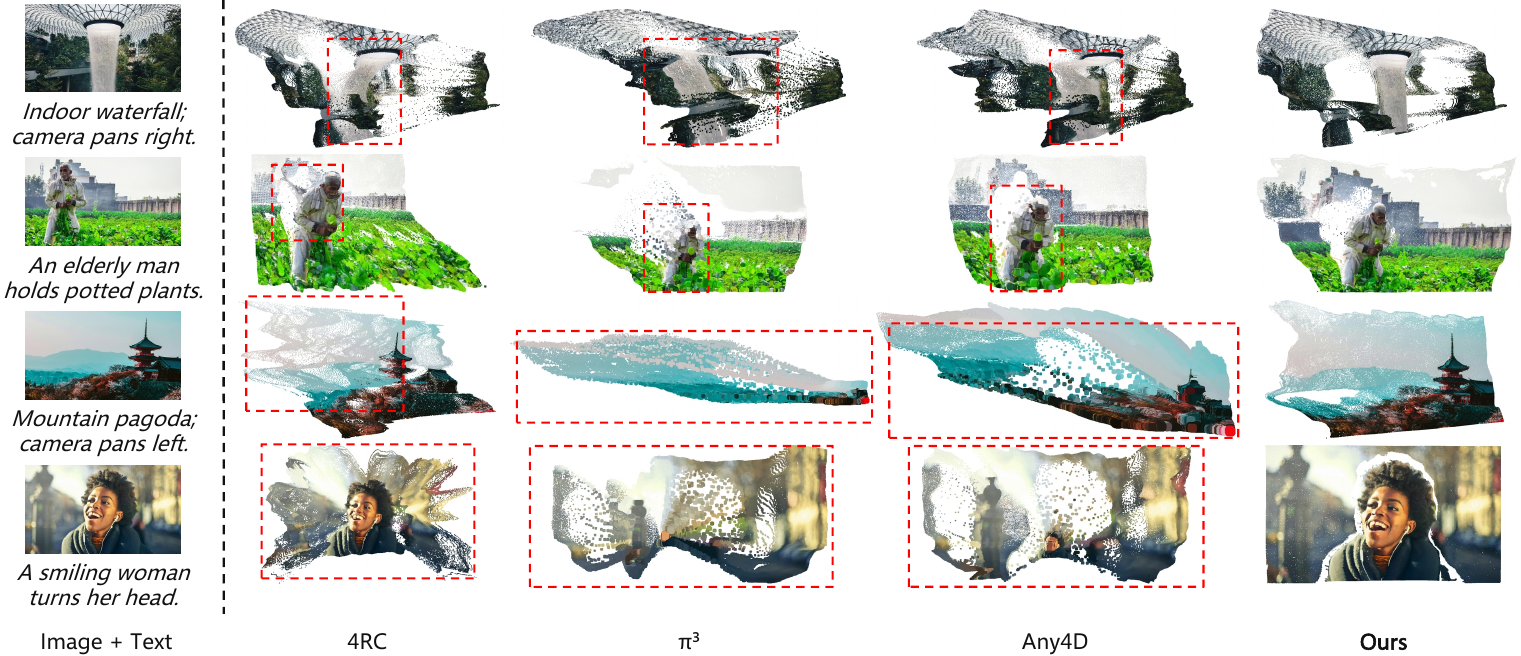}
    \caption{\textbf{Image-to-4D comparison.}
    Each row shows the input condition and 4D results. RGB baselines reconstruct
    the decoded video, whereas Ours consumes its terminal latent directly.}
    \label{fig:image4d_qualitative}
\end{figure*}

\subsection{Experimental Setup}
\label{sec:experimental_setup}

\noindent
\textbf{Implementation Details.}
We use the frozen Wan VAE and initialize the 31-block refinement hierarchy and
prediction heads from 4RC~\cite{dinov2,4rc}. We train the Alignment Module and
prediction heads while adapting the refinement hierarchy with rank-16 LoRA,
keeping the video models and pretrained base weights frozen. Starting from a
4RC-aligned latent adapter, the model undergoes multi-stage geometry-supervised
training, with the final stage using 1,143 clips from six reconstruction
datasets. All benchmarks are held out and use no test-time
adaptation. Full architecture and training details are provided in the Appendix.

\noindent
\textbf{Benchmarks and Metrics.}
Text4D-200 and I4D-200 are locked 200-case text- and image-conditioned
benchmarks; every method is evaluated on every case. We render each predicted
point sequence from two off-axis cameras and report Text CLIP, RGB-reference
CLIP-I, DINO global similarity, valid-patch DINO matching, and DINO set F1.
The corresponding generated RGB is used only for evaluation and coloring, not
as geometry input. The off-axis DINO scores are appearance-dependent proxies
for visible geometric coherence and completeness, not metric 4D accuracy;
multi-view human judgments and ground-truth ablations provide complementary
evidence. Benchmark construction and metric definitions are given in
Appendix.

\begin{table}[t]
\centering
\scriptsize
\setlength{\tabcolsep}{1.8pt}
\renewcommand{\arraystretch}{1.0}
\newcommand{\prefscore}[2]{%
    \shortstack{{#1}\\[-0.4mm]
    {\fontsize{6.0}{6.2}\selectfont [#2]}}%
}

\begin{tabular*}{\columnwidth}{@{\extracolsep{\fill}}lcccc@{}}
\toprule
Task
& \shortstack{Condition\\fidelity $\uparrow$}
& \shortstack{Geometry and\\completeness $\uparrow$}
& \shortstack{Temporal\\stability $\uparrow$}
& \shortstack{Overall\\quality $\uparrow$} \\
\midrule
Text-to-4D
& \prefscore{59.2}{54.1--64.3}
& \prefscore{66.8}{62.0--71.5}
& \prefscore{63.5}{58.4--68.5}
& \prefscore{65.7}{60.8--70.5} \\
Image-to-4D
& \prefscore{66.4}{61.8--70.9}
& \prefscore{72.1}{67.8--76.3}
& \prefscore{68.3}{63.7--72.8}
& \prefscore{70.6}{66.2--74.9} \\
\bottomrule
\end{tabular*}
\caption{\textbf{User preference for Ours over baselines (\%).}}
\label{tab:user_study}
\end{table}

\begin{table}[t]
\centering
\scriptsize
\setlength{\tabcolsep}{1.7pt}
\renewcommand{\arraystretch}{0.88}

\begin{tabular*}{\columnwidth}{@{\extracolsep{\fill}}lrrrrrr@{}}
\toprule
& \multicolumn{3}{c}{7-Scenes (18)}
& \multicolumn{3}{c}{NRGBD (9)} \\
\cmidrule(lr){2-4}\cmidrule(l){5-7}
Variant
& Acc$\downarrow$ & Comp$\downarrow$ & NC$\uparrow$
& Acc$\downarrow$ & Comp$\downarrow$ & NC$\uparrow$ \\
\midrule
w/o Grid
& 3.783 & 6.844  & 0.608
& 5.823 & 9.686  & 0.726 \\
w/o 3D Conv
& 6.944 & 15.806 & 0.554
& 12.439 & 26.511 & 0.594 \\
w/o Frame
& 6.688 & 20.806 & 0.513
& 12.367 & 36.982 & 0.502 \\
w/o Global
& 6.754 & 19.742 & 0.559
& 13.818 & 36.207 & 0.515 \\
\textbf{Full}
& \textbf{3.121} & \textbf{5.418} & \textbf{0.628}
& \textbf{5.202} & \textbf{8.187} & \textbf{0.766} \\
\bottomrule
\end{tabular*}
\caption{\textbf{Component ablation on 7-Scenes and NRGBD.}
Acc/Comp are in cm; lower is better except for NC.}
\label{tab:core_ablations}
\end{table}

\begin{figure}[t]
    \centering
    \vspace{-5pt}
    \includegraphics[width=\linewidth]{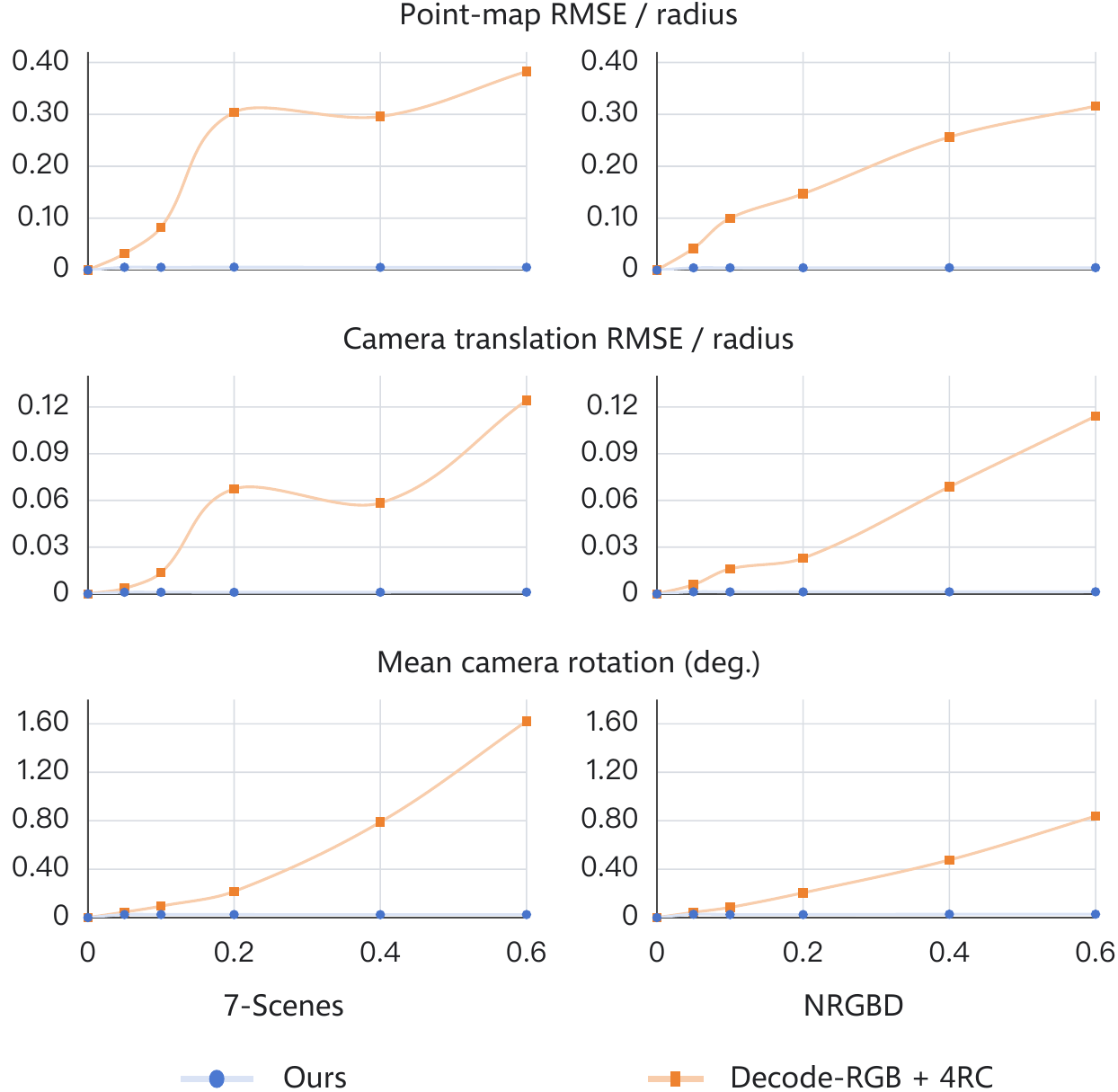}
    \captionof{figure}{\textbf{Grid-Align-null DiT-residual sensitivity.}
    Geometry and camera drift on 7-Scenes and NRGBD.}
    \vspace{-5pt}
    \label{fig:latent_perturbation}
\end{figure}

\noindent
\textbf{Baselines.}
One checkpoint serves Wan2.1-T2V-14B, Wan2.1-T2V-1.3B~\cite{wan}, and
Wan2.2-I2V-A14B, which share the frozen Wan VAE. We compare with generate-then-reconstruct that decode the corresponding latent and apply 4RC~\cite{4rc},
$\pi^3$~\cite{pi3}, or Any4D~\cite{any4d}; CogVideoX-5B~\cite{cogvideox}
provides an additional text-video generator, and official 4DNeX~\cite{4dnex}
provides a native Image-to-4D baseline. Ours versus the matched Wan cascades is
the controlled same-latent comparison. These generated-latent benchmarks are
our primary robustness test: L4AR is trained only on observed-video encodings
but evaluated unchanged on samples from three conditional DiTs. Further
protocol details are in Appendix.

\begin{figure}[t]
    \centering
    \vspace{-3pt}
    \includegraphics[width=\linewidth]{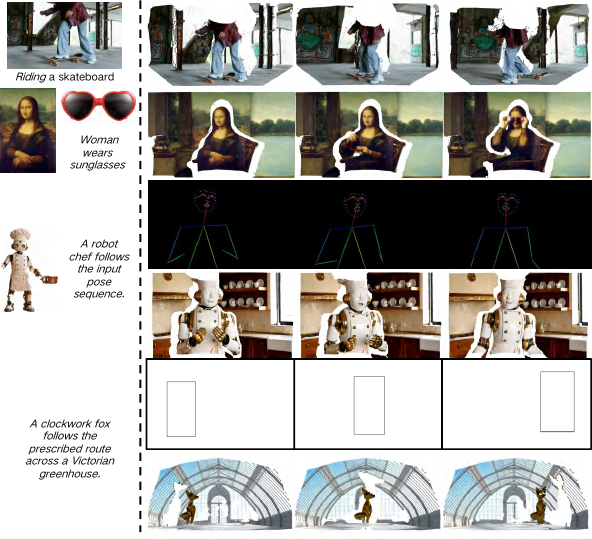}
    \caption{\textbf{Additional controls.}
    Motion, appearance, pose, and trajectory inputs.}
    \label{fig:broader_controls}
\end{figure}

\subsection{Quantitative 4D Generation Results}
\label{sec:quantitative_4d_results}

\noindent
\textbf{Text-to-4D.}
Direct latent lifting consistently outperforms the matched Wan+4RC cascades
and the other tested reconstructors on structure-sensitive DINO metrics, with
DINO-F1 gains of 2.88--3.45 points over matched 4RC. One checkpoint gives
nearly identical performance with both text DiTs, supporting reuse without
retraining. CogVideoX-5B+4RC retains the highest RGB-ref. CLIP-I, so we do not
claim a uniform win on every metric.

\noindent
\textbf{Image-to-4D.}
Ours ranks first on all I4D-200 metrics, including a 5.81-point DINO-F1 gain
over the matched Wan2.2+4RC cascade, and exceeds the alternative
reconstructors and native 4DNeX. Across both tasks, the off-axis DINO gains are
consistent with more complete visible geometry, while not constituting metric
geometry measurements.


Every interval exceeds 50\%, with the strongest preferences for geometry and
completeness. These judgments support more plausible and stable 4D outputs,
but do not measure metric geometry.


\subsection{Qualitative 4D Generation Results}
\label{sec:qualitative_4d_results}

\noindent
\textbf{Text-to-4D.}
Figure~\ref{fig:text4d_qualitative} covers articulated subjects, interacting
elements, thin structures, translucency, and subtle motion.  Ours generally
preserves the principal subject and more surrounding scene support, whereas
the RGB baselines exhibit holes, fragmented surfaces, or missing structures.
The displayed point clouds show one off-axis frame and therefore assess
geometric plausibility, completeness, and content preservation rather than
temporal stability.

\begin{figure}[t]
    \centering
    \vspace{-3pt}
    \includegraphics[width=\linewidth]{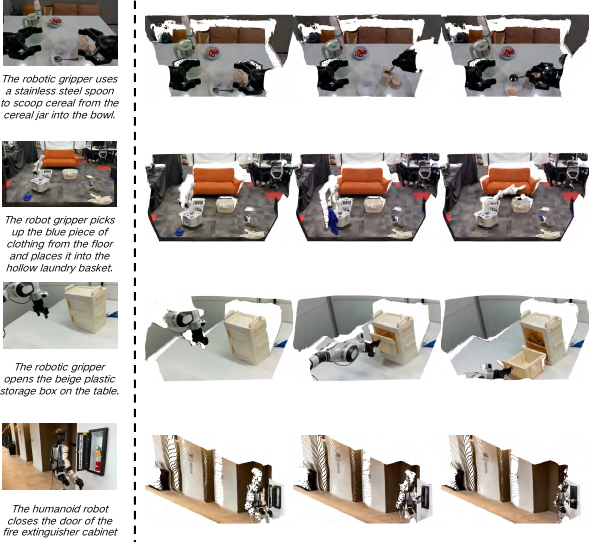}
    \caption{\textbf{Action-conditioned 4D.}
    Manipulation and navigation latents from a compatible backbone.}
    \label{fig:world_action_applications}
\end{figure}

\noindent
\textbf{Image-to-4D.}
Figure~\ref{fig:image4d_qualitative} spans indoor and outdoor scenes, people,
animals, and camera motion. Ours retains more foreground and source structure
than the generate-then-reconstruct, consistent with Table~\ref{tab:generated4d_main}.

\noindent
\textbf{Human Evaluation.}
Fifty participants evaluated 50 sampled cases per benchmark through
randomized, anonymized pairwise comparisons. They could replay motion and
inspect multiple viewpoints before judging condition fidelity, geometry and
completeness, temporal stability, and overall quality. Each case received ten
ratings; Table~\ref{tab:user_study} reports case-averaged preferences with
95\% bootstrap intervals. The full
protocol is provided in Appendix.

\subsection{Diagnostic Analyses}
\label{sec:further_analysis}

\noindent
\textbf{Sensitivity to DiT-Derived Residuals.}
\label{sec:latent_perturbation}
We project the empirical near-terminal residual $\mathbf z_{45}-\mathbf z_{50}$ onto the Grid-Align width-null space and add it to an observed-video latent, applying the same perturbation to both pathways; Appendix provides the construction. All 30 comparisons favor Ours in Fig~\ref{fig:latent_perturbation}. At $\rho=0.6$, point-map drift is $0.0053/0.0047$ for Ours versus $0.3827/0.3160$ for the baseline on 7-Scenes/NRGBD, with the same trend for camera estimation. This isolates a residual component rather than arbitrary DiT errors.

\noindent
\textbf{Benefits of Alignment and Refinement.}
\label{sec:core_ablations}
Table~\ref{tab:core_ablations} uses the matched protocol in
Appendix. Every removal degrades both
ground-truth benchmarks; the largest drops arise without the 3D convolution
or either attention scope. This validates the design rather than
claiming reconstruction OOD superiority.

\subsection{Broader Applications}
\label{sec:broader_applications}

The same L4AR checkpoint inherits upstream motion, appearance, pose,
trajectory, manipulation, and navigation controls
(Figures~\ref{fig:broader_controls}--\ref{fig:world_action_applications}).
These examples show interface compatibility, not action success or physical
correctness.

%% file: sections/5_conclusion.tex
\section{Conclusion}
\label{sec:conclusion}


We introduce direct latent-to-4D generation, using a video model's final
denoised VAE latent as a reusable interface to a 4D decoding hierarchy initialized
from a pretrained reconstructor. Trained on roughly 1K reconstruction
clips, Latent-to-4D transfers one checkpoint unchanged across three compatible
text- and image-conditioned DiTs. Benchmarks yield higher
projection-based DINO-F1 than RGB-decoding cascades, while multi-view
human evaluation favors geometric plausibility, completeness, and temporal
stability. A DiT-residual diagnostic and ground-truth ablations probe
interface sensitivity and support the L4AR design. Qualitative results
demonstrate compatibility with upstream controls without condition-specific
training. Evidence remains limited to a shared VAE convention, and
projection-based evaluation does not establish metric accuracy for generated scenes.